\pdfoutput=1

\documentclass[11pt]{article}

\usepackage[]{ACL2023}

\usepackage{times}
\usepackage{latexsym}

\usepackage[T1]{fontenc}

\usepackage[utf8]{inputenc}

\usepackage{microtype}

\usepackage{inconsolata}

\usepackage[whole]{bxcjkjatype}

\usepackage{subcaption}

\usepackage{multirow}
\usepackage{booktabs}

\usepackage{opensans}

\usepackage{amsmath,amssymb,mathtools}
\usepackage{tikz}
\usepackage{graphicx}

\newcommand{\mathboldface}[1]{\boldsymbol{#1}}
\newcommand{\bm}[1]{\mathboldface{#1}}

\usepackage{fontawesome}

\usepackage[normalem]{ulem}

\title{Transformer Language Models Handle Word Frequency \\in Prediction Head}

\author{
Goro\,Kobayashi$^{1,3}$\hspace{1em}
Tatsuki\,Kuribayashi$^{2,1}$\hspace{1em}
Sho\,Yokoi$^{1,3}$\hspace{1em}
Kentaro\,Inui$^{1,3}$\\[2pt]
$^{1}$ Tohoku University\hspace{1em}
$^{2}$ MBZUAI\hspace{1em}
$^{3}$ RIKEN\hspace{1em}
\\
\texttt{goro.koba@dc.tohoku.ac.jp}\hspace{1em}
\texttt{tatsuki.kuribayashi@mbzuai.ac.ae} \\
\texttt{\{yokoi, kentaro.inui\}@tohoku.ac.jp} \\
}

\begin{document}
\maketitle
\begin{abstract}

Prediction head is a crucial component of Transformer language models.
Despite its direct impact on prediction, this component has often been overlooked in analyzing Transformers.
In this study, we investigate the inner workings of the prediction head, specifically focusing on bias parameters.
Our experiments with BERT and GPT-2 models reveal that the biases in their word prediction heads play a significant role in the models' ability to reflect word frequency in a corpus, aligning with the logit adjustment method commonly used in long-tailed learning. 
We also quantify the effect of controlling the biases in practical auto-regressive text generation scenarios;
under a particular setting, more diverse text can be generated without compromising text quality.

\large{\faicon{github}}
\hspace{.25em}
\parbox{\dimexpr\linewidth-6\fboxsep-2\fboxrule}{\sloppy \small \url{https://github.com/gorokoba560/transformer-lm-word-freq-bias}}

\end{abstract}

\section{Introduction}
\label{sec:intro}
Transformer language models (TLMs)~\cite{devlin2018bert,radford2019gpt2} %
are now fundamental to natural language processing (NLP) techniques, including text generation.
Owing to this success, extensive %
research has been conducted to analyze their inner workings~\cite{rogers-etal-2020-primer-bertology,geva-etal-2021-transformer}.

\begin{figure}[t]
\centering
\centering
\includegraphics[width=\hsize]{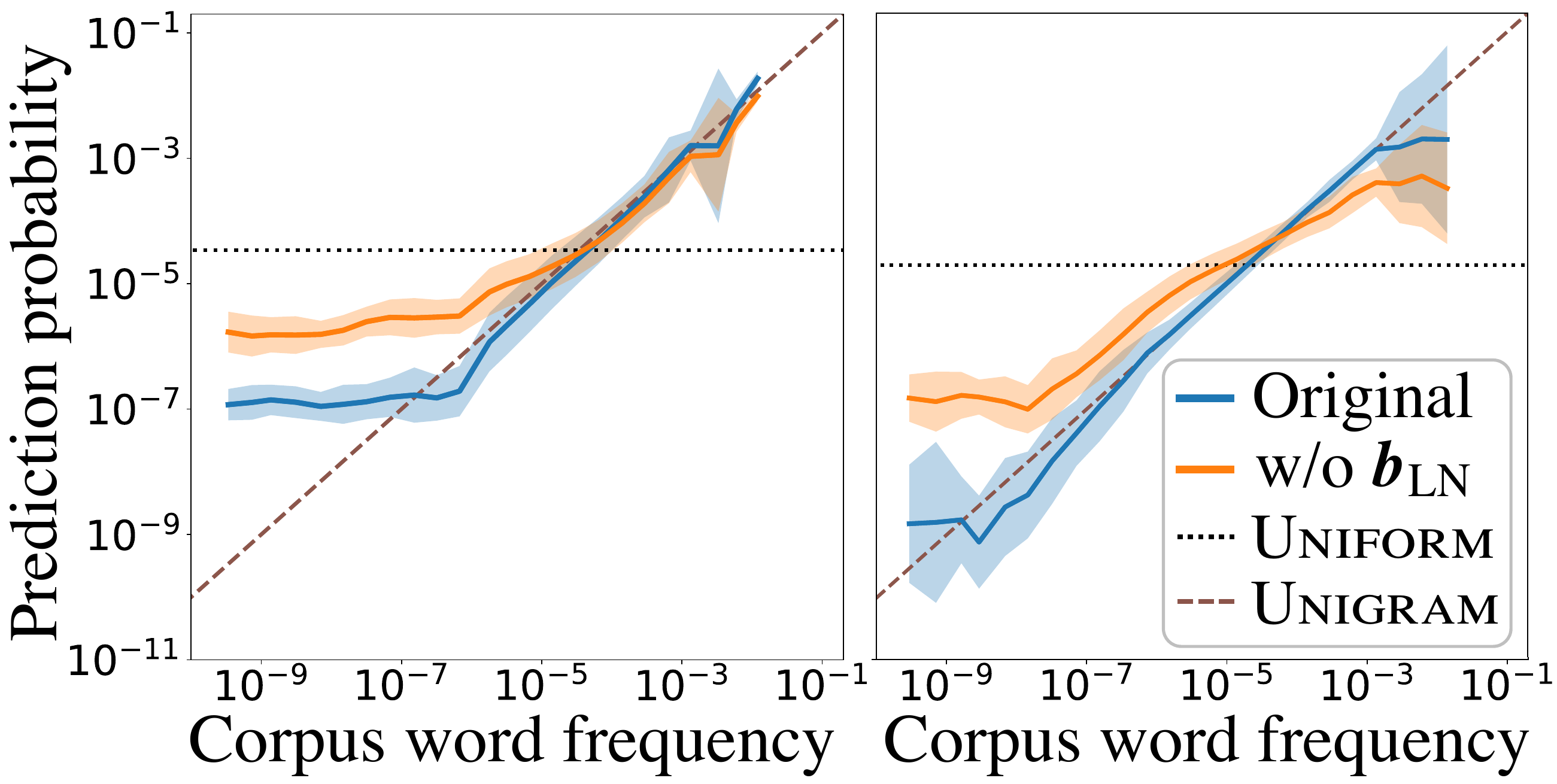}
\caption{
    Changes in word prediction probabilities due to the removal of bias $\bm b_{\mathrm{LN}}$ from BERT base (left) and GPT-2 small (right).
}
\label{fig:bert_base_prob}
\end{figure}

In this study, we shed light on the operation of the \textbf{prediction head}, the last block of the TLMs. %
Despite its direct impact on TLMs' output, its characteristics have been overlooked in previous analyses.
Our experiments with BERT and GPT-2 reveal that \textbf{a particular bias parameter in the prediction head adjusts the model's output toward word frequency in a corpus}. %
Particularly, the bias increases the prediction probability for high-frequency words and vice versa (Figure~\ref{fig:bert_base_prob}).

We further explore this phenomenon from several perspectives.
First, we analyze the geometric characteristics of this phenomenon, which show that word frequency is encoded in a specific direction in the output embedding space. %
Second, we analyze the behavioral impact of controlling their frequency biases on text generation.
The results demonstrate that the model's text generation can be made more diverse while maintaining the fluency by adequately decaying the bias parameters, suggesting that models can more or less isolate word frequency knowledge from other text generation ability. %
Third, we discuss the potential connection between our findings and the logit adjustment method that is typically used in the machine learning field to address the class imbalance problem.

\section{Background: Prediction Head}
\label{sec:preparation}

\begin{figure}[t]
    \centering
    \includegraphics[width=\hsize]{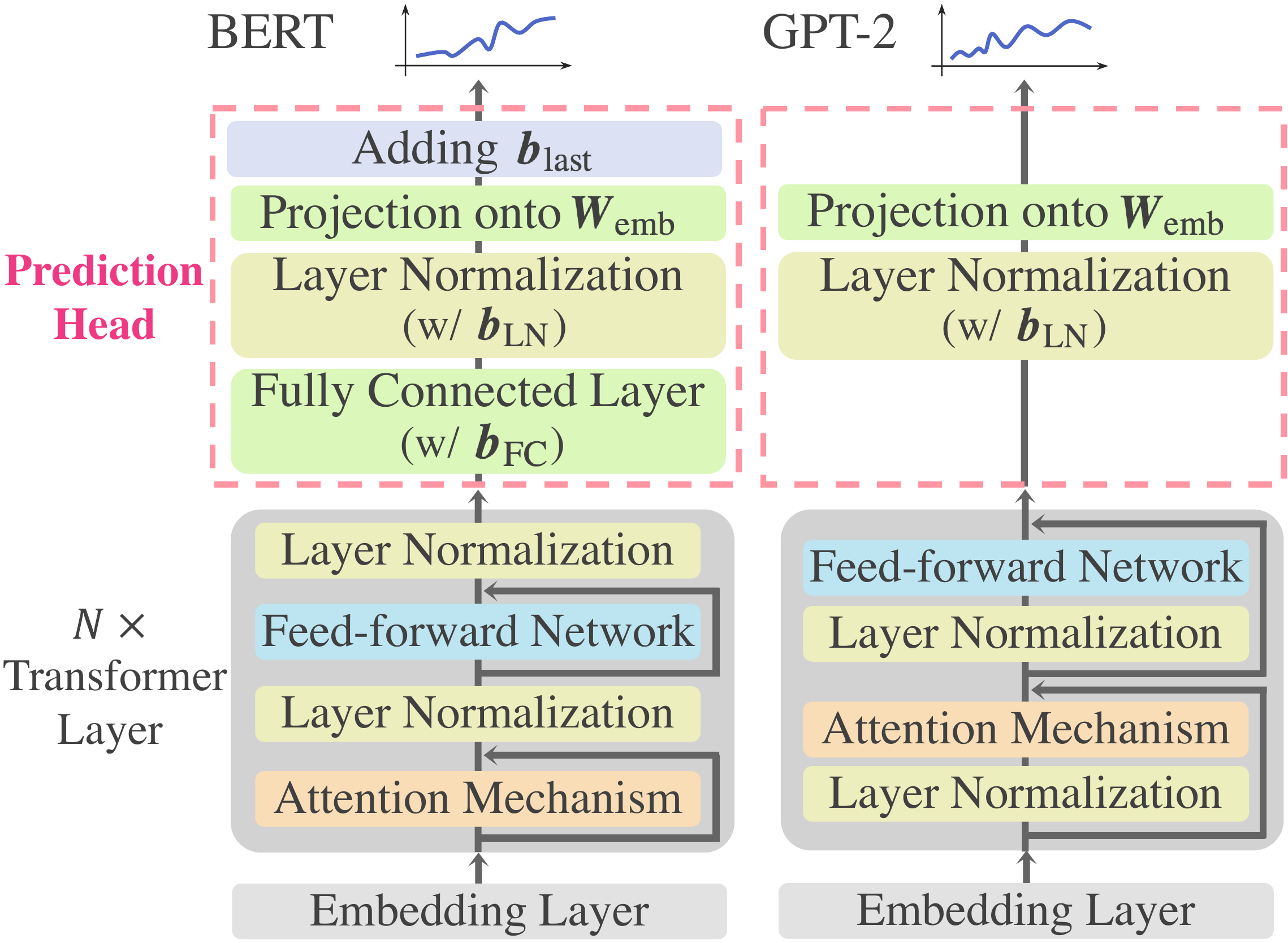}
    \caption{
        Architecture overview of BERT and GPT-2.
    }
    \label{fig:transformer_lm}
\end{figure}

TLMs have a stack of Transformer layers on top of the embedding layer. %
These components update hidden token representations (Figure~\ref{fig:transformer_lm}).
\textbf{Prediction head}, which is our target of analysis, is the last, top-most component in TLMs.
The prediction head in a TLM computes the prediction probabilities for all vocabulary $\mathcal V$ based on the hidden state in the last Transformer layer.

Formally, the prediction head receives, for each token, the last Transformer layer's hidden state $\bm x \in\mathbb{R}^d$. %
The prediction head computes the probability distribution $\bm p \in \mathbb{R}^{\lvert\mathcal V\rvert}$ of the next word as follows, in the case of GPT-2:
\begin{align}
    \bm p &=\mathop{\operatorname{softmax}}\Bigl(\mathrm{LN}(\bm x) \bm W_{\mathrm{emb}}\Bigr) 
    \label{eq:gpt2_p}
    \\
    \mathrm{LN}(\bm x) &\coloneqq \frac{\bm x - m(\bm x)}{s(\bm x)}\odot \bm \gamma + \bm b_{\mathrm{LN}} \in\mathbb{R}^d \text{,}
    \label{eq:ln}
\end{align}
where $\bm W_{\mathrm{emb}}\in\mathbb{R}^{d\times\lvert\mathcal{V}\rvert}$ denotes the word embedding matrix, $m(\bm x)$ and $s(\bm x)$ denote the element-wise mean and standard deviation, respectively,
and $\odot$ denotes the element-wise product.
$\bm \gamma$ and $\bm b_{\mathrm{LN}} \in\mathbb{R}^{d}$ are learnable parameters.

For BERT, there is an additional fully connected layer (FC).
The prediction head computes the probability distribution $\bm p$ for the hidden state $\bm x$ that corresponds to the \texttt{[MASK]} token as follows:
\begin{align}
    \bm p &= \mathop{\operatorname{softmax}}\Bigl(\mathrm{LN}(\bm x^{\prime}) \bm W_{\mathrm{emb}} + \bm b_{\mathrm{last}}\Bigr)
    \label{eq:bert_p}
    \\
    \bm x^{\prime} &= \mathrm{GELU}\bigl(\bm x \bm W_{\mathrm{FC}} + \bm b_{\mathrm{FC}}\bigr) \in\mathbb{R}^d \text{,}
\end{align}
where $\bm W_{\mathrm{FC}}\in\mathbb{R}^{d\times d}$ denotes the learnable weight matrix, and $\bm b_{\mathrm{FC}}\in\mathbb{R}^d$ and $\bm b_{\mathrm{last}}\in\mathbb{R}^{\lvert\mathcal V\rvert}$ denote the learnable bias parameters.
GELU~\cite{hendrycks2016gelu} is the activation function.

Both prediction heads contain the bias $\bm b_{\mathrm{LN}}$, and 
the BERT head
additionally contains the biases $\bm b_{\mathrm{FC}}$ and $\bm b_{\mathrm{last}}$.
As the first step in analyzing the prediction head, we focus on these three biases because they can easily be mapped to the output space.
Drawing on the existing findings about the frequency-related workings of several components in the Transformer~\cite{voita19-bottomup,kobayashi-etal-2020-attention}, we analyze the model behavior with respect to word frequency.

\section{Experiments}
\label{sec:exp}
First, we show that the bias parameters are related to word frequency.
Next, we analyze their properties from two perspectives: (i) geometric characteristics and (ii) text generation.

\paragraph{Model:}
We used BERT (cased)~\cite{devlin2018bert} in two different sizes (base and large) and GPT-2~\cite{radford2019gpt2} in four different sizes (small, medium, large, and xl).

\paragraph{Data:}
We used 5,000 sequences from the test set of the GPT-2 pre-trainng corpus, OpenWebText Corpus~\cite{Gokaslan2019OpenWeb}%
\footnote{
    \texttt{webtext.test.jsonl} published in \url{https://github.com/openai/gpt-2-output-dataset} was used.
}.
Each sequence was fed into BERT after some tokens were replaced with \texttt{[MASK]}%
\footnote{
    Following \citet{devlin2018bert}, 15\% of tokens were replaced with \texttt{[MASK]} 80\% of the time.
}, and fed into GPT-2 as it was.
Further, word frequencies were calculated from the corpus used for training each of BERT and GPT-2.%
\footnote{
    BERT was trained on Wikipedia and BooksCorpus~\cite{Zhu_2015_bookscorpus}, and GPT-2 was trained on OpenWebText Corpus.
    We reproduced them using \texttt{Datasets}~\cite{lhoest2021datasets}.
}

\subsection{Impact of biases on prediction distribution}
\label{sec:exp1:analyze_bias}

We compared the models' word prediction with and without each bias.
Specifically, we once obtained word prediction distributions $\hat{\bm p} \in \mathbb{R}^{\lvert\mathcal V\rvert}$ from a model for each time step across the test data. The average of these distributions are referred to as \textit{model's word prediction distribution} henceforth.

\subsubsection*{Bias adjusts the model's prediction distribution closer to the corpus frequency distribution:}
Figure~\ref{fig:bert_base_prob} shows changes in the model's word prediction distribution before and after the bias $\bm b_{\mathrm{LN}}$ is removed.%
\footnote{
    We created bins to divide the corpus word frequencies into constant intervals and plotted each bin's geometric mean and standard deviation for the word prediction probabilities.
}
The removal of $\bm b_{\mathrm{LN}}$ increases the probability of the model predicting low-frequency words (right side of the figures) and vice versa, which results in a word prediction distribution that approaches a flat (\textsc{Uniform} in the figure).
In other words, the bias $\bm b_{\mathrm{LN}}$ adjusts the models' word prediction distribution to be closer to the corpus word frequency distribution (\textsc{Unigram} in the figure).
This finding can be generalized across all model sizes (Appendix~\ref{ap:results_on_other_sizes}).

\begin{table}[t]
\centering
\small
\setlength{\tabcolsep}{3.5pt}  %
\begin{tabular}{@{}llcccc@{}}
\toprule
\multicolumn{2}{c}{Model} & \multicolumn{1}{c}{Original} 
& \multicolumn{1}{c}{w/o $\bm b_{\mathrm{LN}}$}
& \multicolumn{1}{c}{w/o $\bm b_{\mathrm{FC}}$}
& \multicolumn{1}{c}{w/o $\bm b_{\mathrm{last}}$} \\ 
\cmidrule{1-6}
\morecmidrules
\cmidrule{1-6}
\multirow{2}{*}{BERT}
& base   & 0.20 & \textbf{0.39} & 0.22 & 0.23 \\
& large  & 0.21 & \textbf{0.39} & 0.23 & 0.23 \\
\cmidrule{1-6}
\multirow{4}{*}{GPT-2}
& small  & 0.14 & \textbf{0.83} &  -    &  -   \\
& medium & 0.14 & \textbf{0.34} &  -    &  -   \\
& large  & 0.14 & \textbf{0.17} &  -    &  -    \\
& xl     & 0.14 & \textbf{0.17} &  -    &  -  \\ 
\bottomrule
\end{tabular}
\caption{
    KL divergence between the model's word prediction distribution and the corpus word frequency distribution.
    A larger value means that the distributions are more divergent. %
    $\bm b_{\mathrm{FC}}$ and $\bm b_{\mathrm{bias}}$ are contained only in BERT.
}
\label{tab:prob_kl_div}
\end{table}

To quantify the above effect, we calculated the Kullback--Leibler (KL) divergence between the model's word prediction distribution and the corpus word frequency distribution (\textsc{Unigram}).
Note that a \textit{higher} value indicates that the model's prediction distribution has \textit{more discrepancy} with that in the pretraining corpus.
Table~\ref{tab:prob_kl_div} shows that removing $\bm b_{\mathrm{LN}}$ always results in a higher value, %
which indicates that $\bm b_{\mathrm{LN}}$ indeed adjusts the prediction distribution to be closer to the corpus frequency distribution.
The biases $\bm b_{\mathrm{FC}}$ and $\bm b_{\mathrm{last}}$ in BERT also exert a similar effect, but it is weaker than that of $\bm b_{\mathrm{LN}}$; we focus on $\bm b_{\mathrm{LN}}$ in the following. %
We also observe that larger models have less change of frequency biases due to $\bm b_{\mathrm{LN}}$.

\subsection{Geometric observations}
\label{sec:exp:geometrical_observation}
We observed the geometric properties of the bias $\bm b_{\mathrm{LN}}$ and the output embedding space of the TLMs.

\subsubsection*{Word frequency is encoded in the bias vector's direction in the output embedding space:}

\begin{figure}[t]
\centering
\includegraphics[width=\hsize]{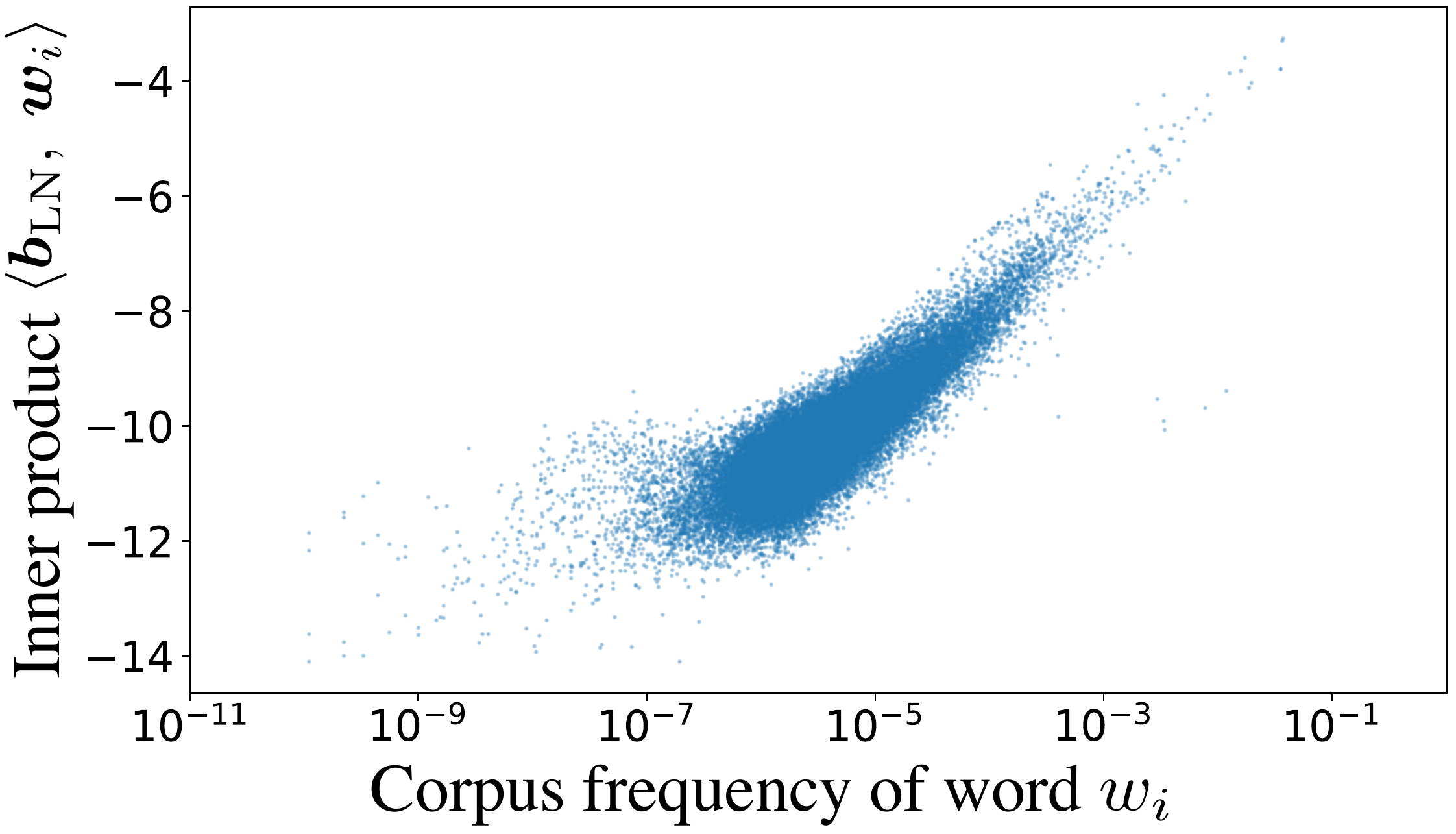}
\caption{
    Relationship between the corpus word frequency and inner product of $\bm b_{\mathrm{LN}}$ and each output word embedding $\bm w_i$ in GPT-2 small.
}
\label{fig:gpt2_bias_parameter}
\end{figure}

The observation that the bias vector shifts predictions according to word frequency
suggests that word frequency is encoded in the output embedding space $\bm W_{\mathrm{emb}}$, and the bias vector $\bm b_{\mathrm{LN}}$ is a good projection to extract this frequency information.
In fact, the inner product of $\bm b_{\mathrm{LN}}$ and each %
word embedding $\bm w_i$ in the embedding layer correlates well with the word frequency%
\footnote{
    Spearman's $\rho$ was $0.78$ on GPT-2 small.
}
(Figure~\ref{fig:gpt2_bias_parameter}).

Furthermore, we observed that removing the bias direction ($\approx$ frequency direction) from the embedding matrix $\bm W_{\mathrm{emb}}$ improved the isotropy (uniformity in direction, e.g.,~\citealp[]{ethayarajh-2019-contextual}) in the output embedding space.
Formally, we removed the bias direction using $\bm w_i \gets \bm w_i - \langle \bm w_i, \frac{\bm b_{\mathrm{LN}}}{\lVert \bm b_{\mathrm{LN}} \rVert}\rangle \frac{\bm b_{\mathrm{LN}}}{\lVert \bm b_{\mathrm{LN}} \rVert}$; 
then, the average value $\frac{1}{n^2}\sum_i \sum_j \cos(\bm w_i, \bm w_j)$ decreased from $0.15$ to $0.09$ in BERT base.
This observation shows that the anisotropy in the output space is, more or less, caused by the frequency direction. %

We further observed that hidden states $\bm h_{\mathrm{token}}$ before $\bm b_{\mathrm{LN}}$ was added were almost orthogonal to $\bm b_{\mathrm{LN}}$ ($\approx$ word frequency direction); in particular, $\mathbb{E}_{\mathrm{token}} \left\lvert \cos\left(\bm h_{\mathrm{token}}, \; \bm b_{\mathrm{LN}}\right)\right\rvert = 0.08 \ll 1.0$ in BERT-base.
This corroborates that the frequency bias injected in the prediction head indeed does not exist in the hidden states before the prediction head.

\subsubsection*{Word frequency encoded on the bias vector is shifted via fine-tuning:}
We also inspected whether the model's word prediction distribution is shifted to that in the target domain during fine-tuning to enhance the generality of our observation.
Specifically, we fine-tuned GPT-2 small on a dataset consisting of abstracts from papers in the machine learning field%
\footnote{
    \texttt{CShorten/ML-ArXiv-Papers} published in \url{https://huggingface.co/datasets/CShorten/ML-ArXiv-Papers} on
    \texttt{Datasets}~\cite{lhoest2021datasets} was used.
}, whose word frequency distribution is different from the pretraining data.
After fine-tuning, the inner product of $\bm b_{\mathrm{LN}}$ and each %
word embedding $\bm w_i$ correlated more with the additional fine-tuning corpus after fine-tuning (the Spearman's $\rho$ changed from $0.38$ to $0.62$) and slightly less with the pre-training corpus (the Spearman's $\rho$ changed from $0.78$ to $0.73$).
This suggests that frequency information captured by the bias $\bm b_{\mathrm{LN}}$ is updated during fine-tuning.

\subsection{Impact of bias on text generation}
\label{sec:exp2:control_bias}

We next demonstrate that controlling the bias $\bm b_{\mathrm{LN}}$ can lead to more diverse text generation without significant harm to the quality of the text. %
We hope that quantifying the effect of such control using metrics for the evaluation of text generation (e.g., n-gram diversity) will enhance the connection %
between the language generation field and the field of probing/interpreting LMs' internals.

\paragraph{Procedure:}
We adjusted $\bm b_{\mathrm{LN}}$ during text generation by GPT-2, and we then evaluated the generated text. %
Specifically, we introduced an adjustment coefficient $\lambda\in[0, 1]$ and replaced $\bm b_{\mathrm{LN}}$ with $\lambda \bm b_{\mathrm{LN}}$.
We report the evaluation scores by varying $\lambda$. %
The results generated with the top-p sampling strategy~\cite{fan-etal-2018-hierarchical} are reported in this section. 
The results for other decoding settings are in Appendix~\ref{ap:results_on_other_sizes}; we found similar results for the top-p and top-k sampling but found degradation with the vanilla sampling setting.
The details of the settings are described in Appendix~\ref{ap:generate_detail}.

\paragraph{Evaluation methods:}
Text generated by each model was evaluated from two perspectives: diversity and quality.
For the diversity evaluation, Distinct-n $D_n$~\cite{li-etal-2016-distinct-n} and N-gram diversity $D$~\cite{meister2022diversity} were used.
These measures of $n$-gram overlap in generated texts were calculated as follows:
\begin{align}
    D_n(\text{texts}) &\coloneqq \frac{\#\;{\text{Unique $n$-grams in texts}}}{\#\;{\text{$n$-grams in texts}}}
    \\
    D(\text{texts}) &\coloneqq \frac{1}{4}\sum_{n=1}^4 D_n(\text{texts})
    \text{.}
\end{align}
For the quality evaluation, \textsc{Mauve}~\cite{pillutla2021mauve} and Perplexity (\textsc{Ppl}) were used.
\textsc{Mauve} evaluates how similar a given text generation model is to humans by comparing human-written texts and model-generated texts according to the difference in their distributions in a sentence embedding space.
\textsc{Ppl} evaluates how well models can predict words in human-written texts. %
\begin{table}[t]
\centering
\small
\setlength{\tabcolsep}{5pt}
\begin{tabular}{@{}ccccccc@{}}
\toprule
\multicolumn{1}{c}{\multirow{2}{*}{Model}} & \multicolumn{1}{c}{\multirow{2}{*}{\hspace*{1.5mm}$\lambda$\hspace*{1.5mm}}} & \multicolumn{3}{c}{Diversity ↑} & \multicolumn{2}{c}{Quality} \\ 
\cmidrule(lr){3-5} \cmidrule(l){6-7} 
\multicolumn{1}{c}{} 
& \multicolumn{1}{c}{} 
& \multicolumn{1}{c}{$D_1$} 
& \multicolumn{1}{c}{$D_2$} 
& \multicolumn{1}{c}{$D$} 
& \multicolumn{1}{c}{\textsc{Mauve} ↑} 
& \multicolumn{1}{c}{\textsc{Ppl} ↓} \\ 
\cmidrule{1-7}
\morecmidrules
\cmidrule{1-7}
\multirow{3}{*}{small} 
 & $1$  &  0.04  & 0.32  & 0.49  & \textbf{0.85}  & \textbf{19.4} \\
 & $0.6$ & \textbf{0.06} & \textbf{0.61} & 0.59 & 0.18 &  24.1 \\
 & $0$  & 0.04 & 0.36 & 0.32 & 0.01 & 65.9   \\ 
\cmidrule{1-7}
\multirow{4}{*}{med.}
 & $1$  & 0.05 & 0.35 & 0.51 & 0.90 & \textbf{14.6}  \\
 & $0.9$ & 0.05 & 0.39 & 0.54 & \textbf{0.90} & 14.8 \\
 & $0.2$ & 0.07 &  \textbf{0.63} & 0.60 & 0.14 & 18.8  \\
& $0$  & \textbf{0.08} & 0.60 & 0.55 & 0.06 & 21.3   \\
\cmidrule{1-7}
\multirow{3}{*}{large}
 & $1$  &  0.04  & 0.30  &  0.47   &  0.90 &  \textbf{12.7}   \\
 & $0.5$ & 0.04 & 0.36 & 0.50 & \textbf{0.91} & 12.9  \\
 & $0$  & \textbf{0.04} & \textbf{0.42}  & \textbf{0.54} & 0.86 & 13.6 \\ 
\cmidrule{1-7}
\multirow{3}{*}{xl}
 & $1$  &  0.04  &  0.30 &  0.47  & 0.90  &  \textbf{11.4}   \\
 & $0.7$ & 0.04 & 0.34 & 0.49 & \textbf{0.92} &  11.5 \\
 & $0$  & \textbf{0.04} & \textbf{0.41}  & \textbf{0.53} & 0.86 & 12.1  \\ 
\bottomrule
\end{tabular}
\caption{
    Evaluation results for GPT-2 (top-p sampling) while bias $\bm b_{\mathrm{LN}}$ was controlled with $\lambda$.
    Results for $\lambda=0,1$, and other notable values are listed.
}
\label{tab:generated_text_eval_topp}
\end{table}

\paragraph{Results:}
Table~\ref{tab:generated_text_eval_topp} shows the results.
Weakening the bias $\bm b_{\mathrm{LN}}$ ($\lambda<1$) increased the diversity of the generated text but decreased the \textsc{Ppl} score, exhibiting a general trade-off between them.
Nevertheless, for the larger models, GPT-2 large ($\lambda=0.5$) and xl ($\lambda=0.7$), there was a sweet spot, where the diversity and the \textsc{Mauve} score improved with little %
decrease in \textsc{Ppl}.
This observation can be interpreted as follows. 
The larger models were able to predict the context-dependent probability of low-frequency words as precisely as that of high-frequency words, so promoting low-frequency words with those models improved the diversity while maintaining the quality of the text. 
The smaller models were equally accurate in predicting the probability of high-frequency words but tended to be inaccurate for low-frequency words, so promoting low-frequency words degraded %
the quality of the text.
This interpretation is also consistent with the class imbalance problem, which will be discussed in Section~\ref{subsec:related:logit_adjustment}.
From the application perspective, this observation also suggests that the lexical diversity in text generation can be improved simply by modifying particular parameters in the prediction head. %

We also show several samples of the generated texts (Appendix~\ref{ap:examples_generated_text}).
We generally observed that overly decreasing $\lambda$ incurs (i) more proper nouns, (ii) more repetitions of the same words or similar phrases, and (iii) the generation of ungrammatical text, especially for the small models.
This may also be related to the suppression of the punctuation and end-of-sequence token, which are highly frequent.

\section{Discussion}
\label{sec:discussion}

\subsection{Connection with logit adjustment methods}
\label{subsec:related:logit_adjustment}
We revealed that
adding the bias $\bm b_{\mathrm{LN}}$ (which was performed immediately before the logit was computed)
encourages TLMs to generate high-frequency words, and de-biasing promotes diversity.
This can also be seen as analogous to logit adjustment, which is a common technique for addressing the class imbalance problem, where the label (the word in text generation) frequency distribution is long-tailed~\cite{provost2000imbalanced,zhou-liu-2006-imbalance,collel2016multiclass-imbalanced,menon2021logit-adjustment}.
In particular, \citet{menon2021logit-adjustment} proposed to minimize the balanced error (i.e., an average of per-class errors) by directly adding the label frequency distribution to logits during training but not during inference.
One can find an analogy between the modification of $\bm b_{\mathrm{LN}}$ and their method:
(i) adding the frequency-shifting bias $\bm b_{\mathrm{LN}}$ %
corresponds to the operation of adding the class-frequency-based margins to the logits;
(ii) promoting low-frequency words by removing $\bm b_{\mathrm{LN}}$ during inference corresponds to the way logit adjustment encourages low-class prediction.
In other words, interestingly, TLMs seem to implicitly learn something similar to balanced error minimization without being explicitly designed to do so (e.g., loss function).

\subsection{Connection with a technique to initialize bias parameter with class frequency}
In training neural classification models, using class frequency to initialize the last bias to be added to the logit is a well-known and efficient technique~\cite{trainingNNblog}.
Therefore, our observation that the bias vector at the prediction head (i.e., the last block) encodes word frequency might seem somewhat obvious. %
However, our experimental results showed peculiar trends that might be stemmed from the inductive bias of TLMs.
First, although the initialization technique implies the relationship between the \textit{last} bias $\bm b_{\mathrm{last}}$ and the corpus word frequency, we found that the bias $\bm b_{\mathrm{LN}}\in\mathbb{R}^{d}$, which is further away from the output and less expressive than $\bm b_{\mathrm{last}}\in\mathbb{R}^{\lvert\mathcal V\rvert}$, plays the role in encoding the frequency in BERT (Table~\ref{tab:prob_kl_div}).
For GPT-2, not even $\bm b_{\mathrm{last}}$ exists. 
Second, even $\bm b_{\mathrm{LN}}$ plays a weak role in encoding the frequency in larger models (Table~\ref{tab:prob_kl_div}).
These findings suggest that neural models dynamically determines the role of each internal module according to various factors such as parameter size and architecture.
When and under what conditions the short vector $\bm b_{\mathrm{LN}}$ strongly encodes the frequency is an interesting question and left to future research.

\section{Related work}
\label{sec:related_works}

Transformer layers (e.g., attention patterns) have been the major focus of TLM analysis~\cite{clark19,marecek19balustrades,kobayashi-etal-2021-incorporating,dai-etal-2022-knowledgeneuron}.
The first embedding layer, especially positional encoding, has also been studied~\cite{wang2021positionbert,kiyono-etal-2021-shape}.
This study sheds light on the prediction head, the last block of a TLM, and provides new insights into the working mechanisms of TLMs.

Notably, previous studies have reported that words having a similar frequency are clustered in the embedding spaces of various deep NLP models~\cite{mu2018allbutthetop,chengyue2018frage,provilkov-etal-2020-bpedrop,liang2021isotropic};
our observation agrees with theirs.
In addition to this, we newly discovered that a particular bias parameter in the TLM prediction head corresponds to ``word frequency direction''
in the word embedding space.

\section{Conclusions}
In this study, we explored the workings of bias parameters in the prediction head of TLMs.
Our experiments with BERT and GPT-2 showed that the biases adjust the model's prediction with respect to word frequency.
We further explored this phenomenon and provided the following insights: 
(i) word frequency is encoded in a specific direction (the bias direction) in the output embedding space, 
(ii) properly controlling the bias's effect can encourage more diverse %
language generation without compromising quality, 
and (iii) TLMs are implicitly trained to be potentially consistent with the logit adjustment method.
In future work, we will analyze larger TLMs, e.g., Open Pre-trained Transformers~\cite{zhang2022opt}.
Further, we will analyze the weight parameters in the prediction head in addition to the bias parameters.

\section*{Limitations}
There are mainly two limitations in this study.
First, we still do not consider components other than the bias parameters in the prediction head.
For example, the weight parameters of the prediction head, i.e., $\bm \gamma$ and $\bm W_{\mathrm{FC}}$, can also affect a model's prediction.
Second, our findings do not cover the Transformer language models other than BERT (base and large) and GPT-2 (small, medium, large, and xl).
Consistent findings were obtained for the two main architectures (i.e., encoder-based masked, and decoder-based causal language models) and for various model sizes, although future research is needed to show whether the findings can be generalized to RoBERTa~\cite{liu19}, Open Pre-trained Transformer Language Models (OPT,~\citealp{zhang2022opt}), and other variants.
Considering Transformer encoder-decoder models, such as neural machine translation models and T5~\cite{raffel2020t5}, would also be an interesting future direction.

\section*{Ethics Statement}
This paper sheds light on the workings of the prediction head of the fundamental models in NLP.
In recent years, unintended biases (e.g., gender bias) in neural network models have been problematic.
This paper may help in this direction by encouraging researchers to analyze the prediction head as well as Transformer layers.

\section*{Acknowledgements}
We would like to thank the members of the Tohoku NLP Group for their insightful comments, particularly Shiki Sato for his valuable suggestions regarding experimental settings. 
This work was supported by JSPS KAKENHI Grant Number JP22J21492, JP22H05106; JST CREST Grant Number JPMJCR20D2, Japan; and JST ACT-X Grant Number JPMJAX200S, Japan.

\bibliography{custom}
\bibliographystyle{acl_natbib}

\appendix
\section{Experimental results in other settings}
\label{ap:results_on_other_sizes}

\begin{figure}[ht]
\centering
\includegraphics[width=\hsize]{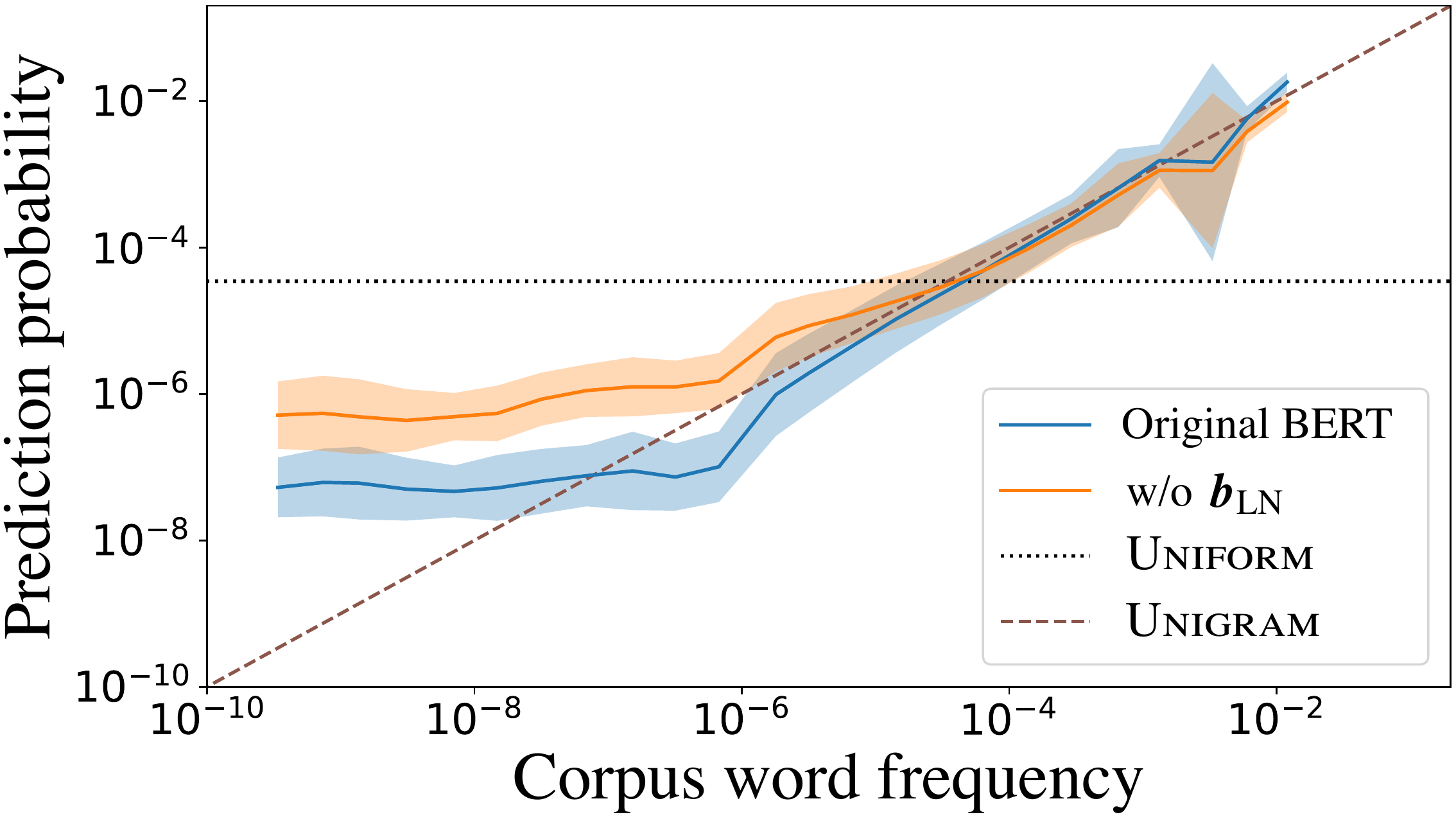}
\caption{
    Changes in word prediction probabilities due to bias $\bm b_{\mathrm{LN}}$ removal on BERT large.
}
\label{fig:bert_large_prob}
\end{figure}

\begin{figure}[ht]
\centering
\includegraphics[width=\hsize]{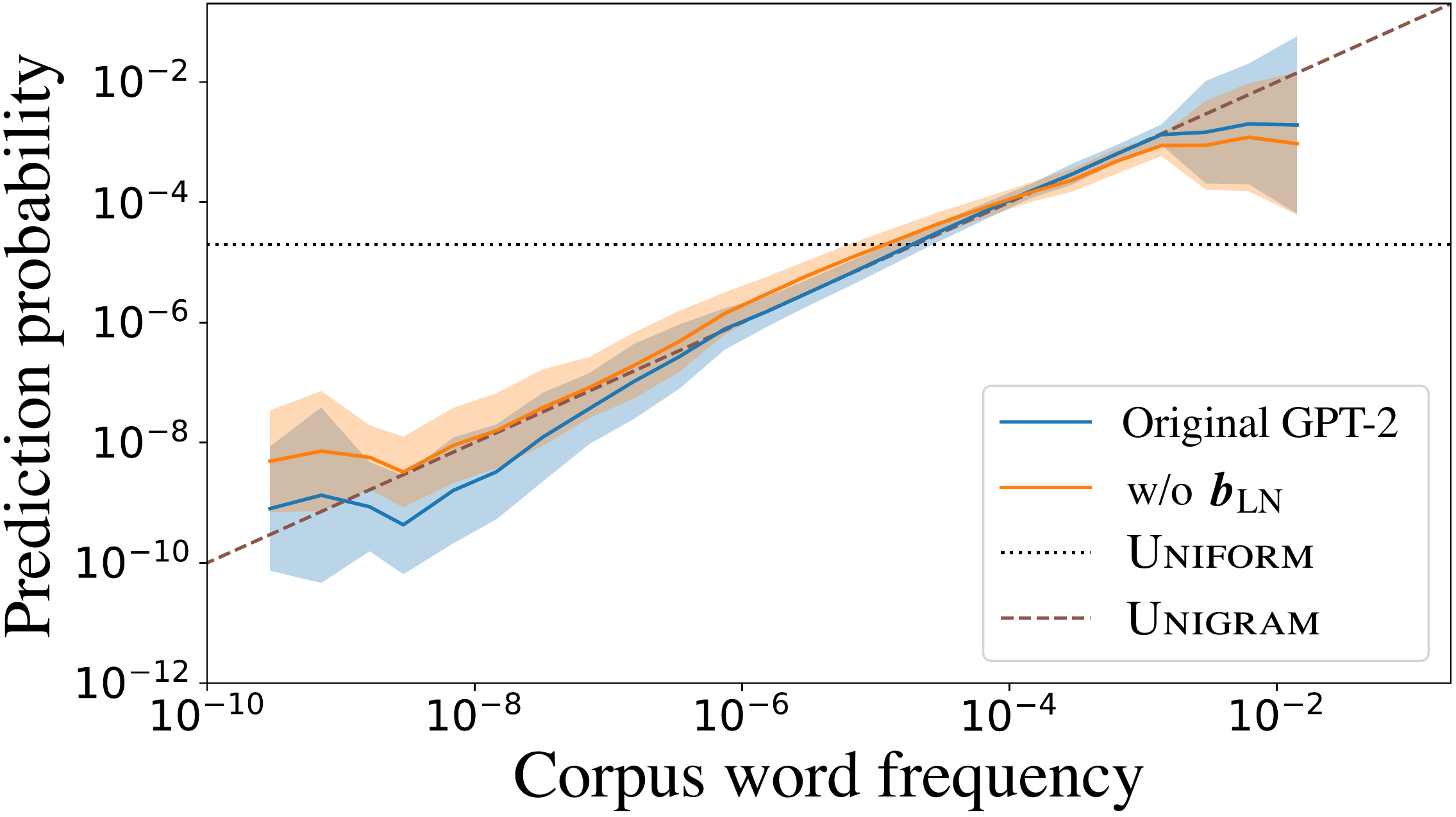}
\caption{
    Changes in word prediction probabilities due to bias $\bm b_{\mathrm{LN}}$ removal on GPT-2 medium.
}
\label{fig:gpt2_medium_prob}
\end{figure}

\begin{figure}[ht]
\centering
\includegraphics[width=\hsize]{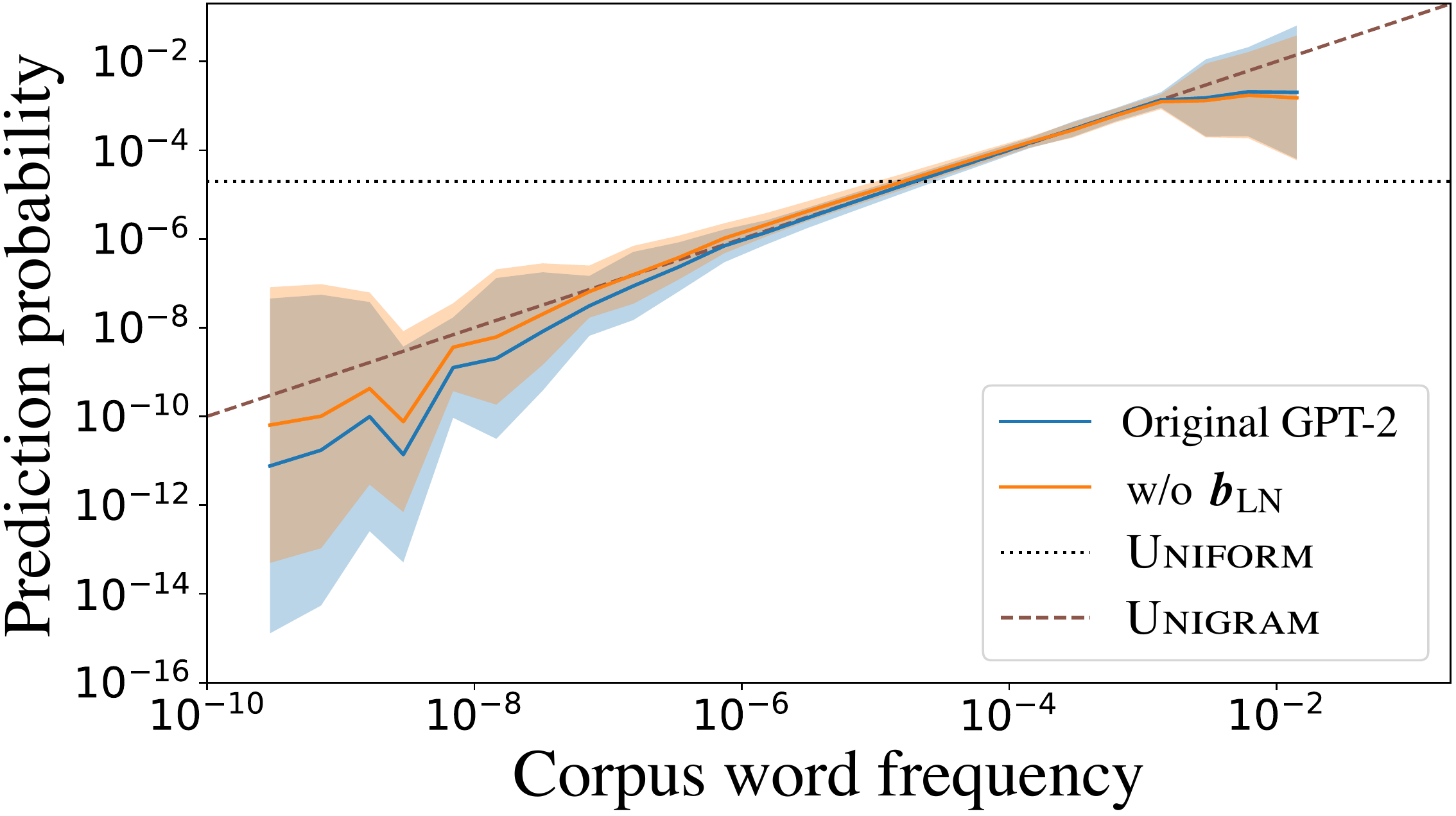}
\caption{
    Changes in word prediction probabilities due to bias $\bm b_{\mathrm{LN}}$ removal on GPT-2 large.
}
\label{fig:gpt2_large_prob}
\end{figure}

\begin{figure}[ht]
\centering
\includegraphics[width=\hsize]{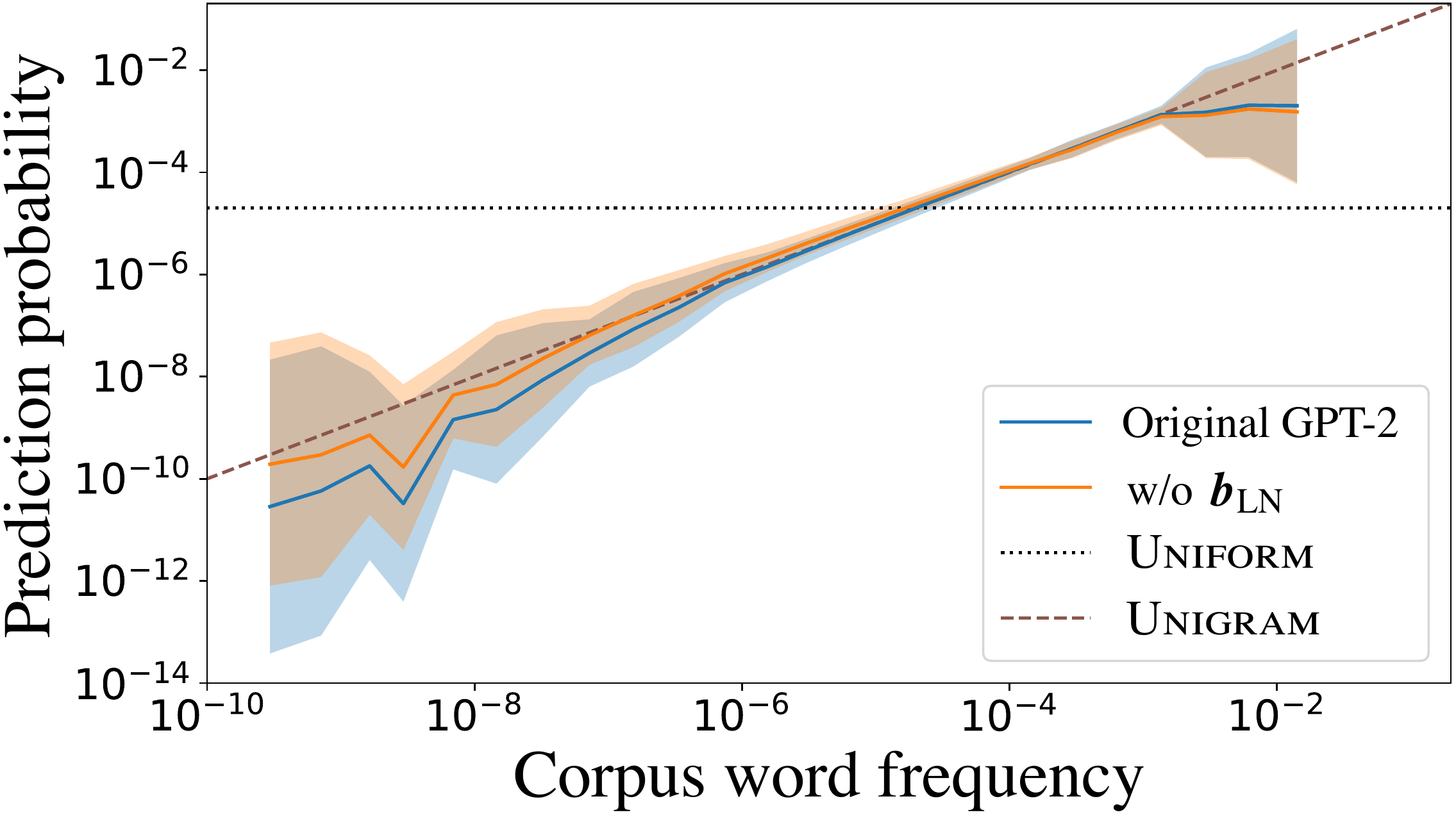}
\caption{
    Changes in word prediction probabilities due to bias $\bm b_{\mathrm{LN}}$ removal on GPT-2 xl.
}
\label{fig:gpt2_xl_prob}
\end{figure}

In Section~\ref{sec:exp1:analyze_bias}, we presented the changes in the word prediction distribution before and after removing the bias $\bm b_{\mathrm{LN}}$ of BERT base and GPT-2 small in Figure~\ref{fig:bert_base_prob}.
The results of the other models are shown in Figures~\ref{fig:bert_large_prob} to \ref{fig:gpt2_xl_prob}.

\begin{figure}[ht]
\centering
\includegraphics[width=\hsize]{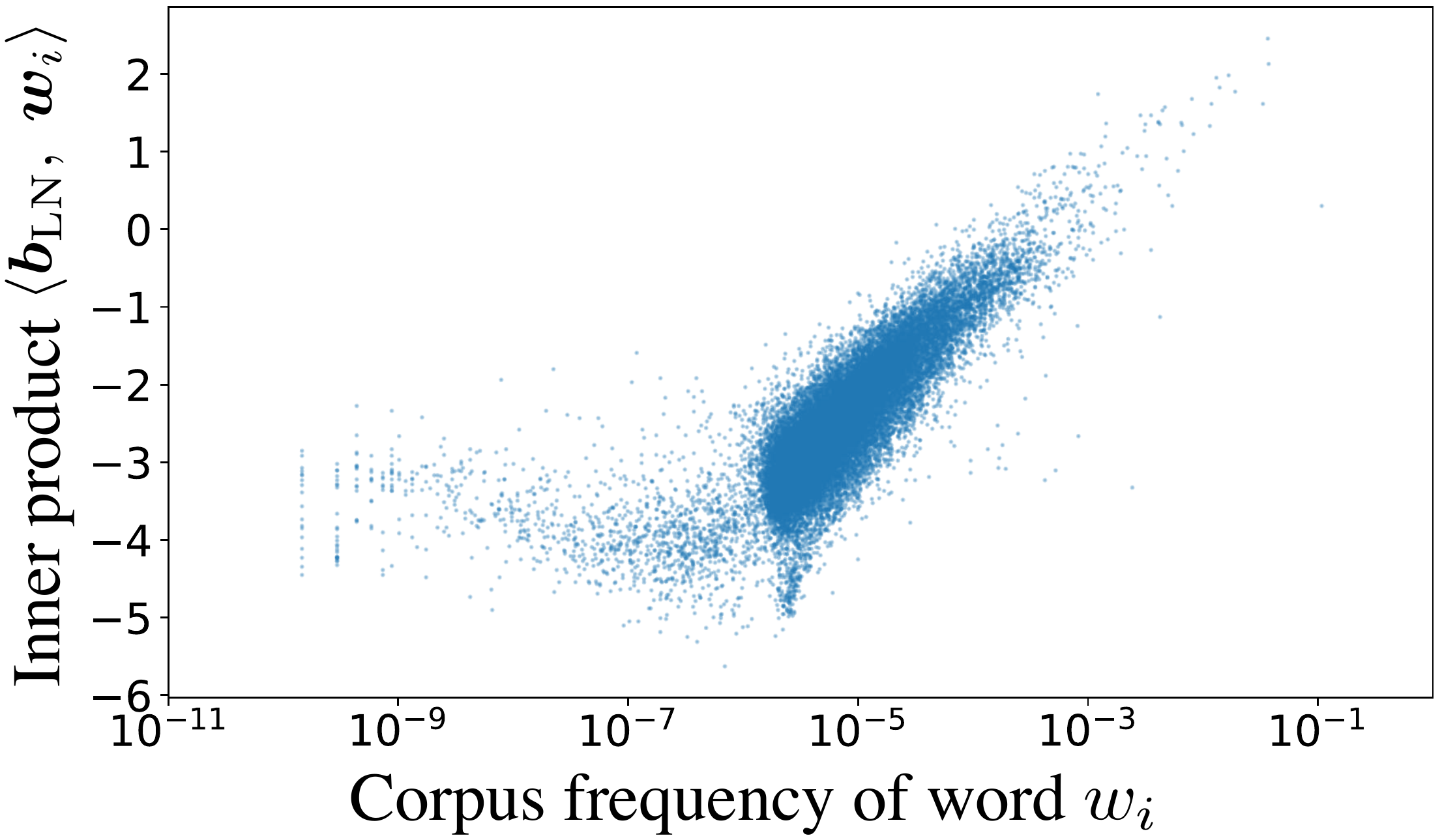}
\caption{
    Relationship between the corpus word frequency and the inner product of $\bm b_{\mathrm{LN}}$ and each output word embedding $\bm w_i$ in BERT base.
}
\label{fig:bert_base_bias_parameter}
\end{figure}

\begin{figure}[ht]
\centering
\includegraphics[width=\hsize]{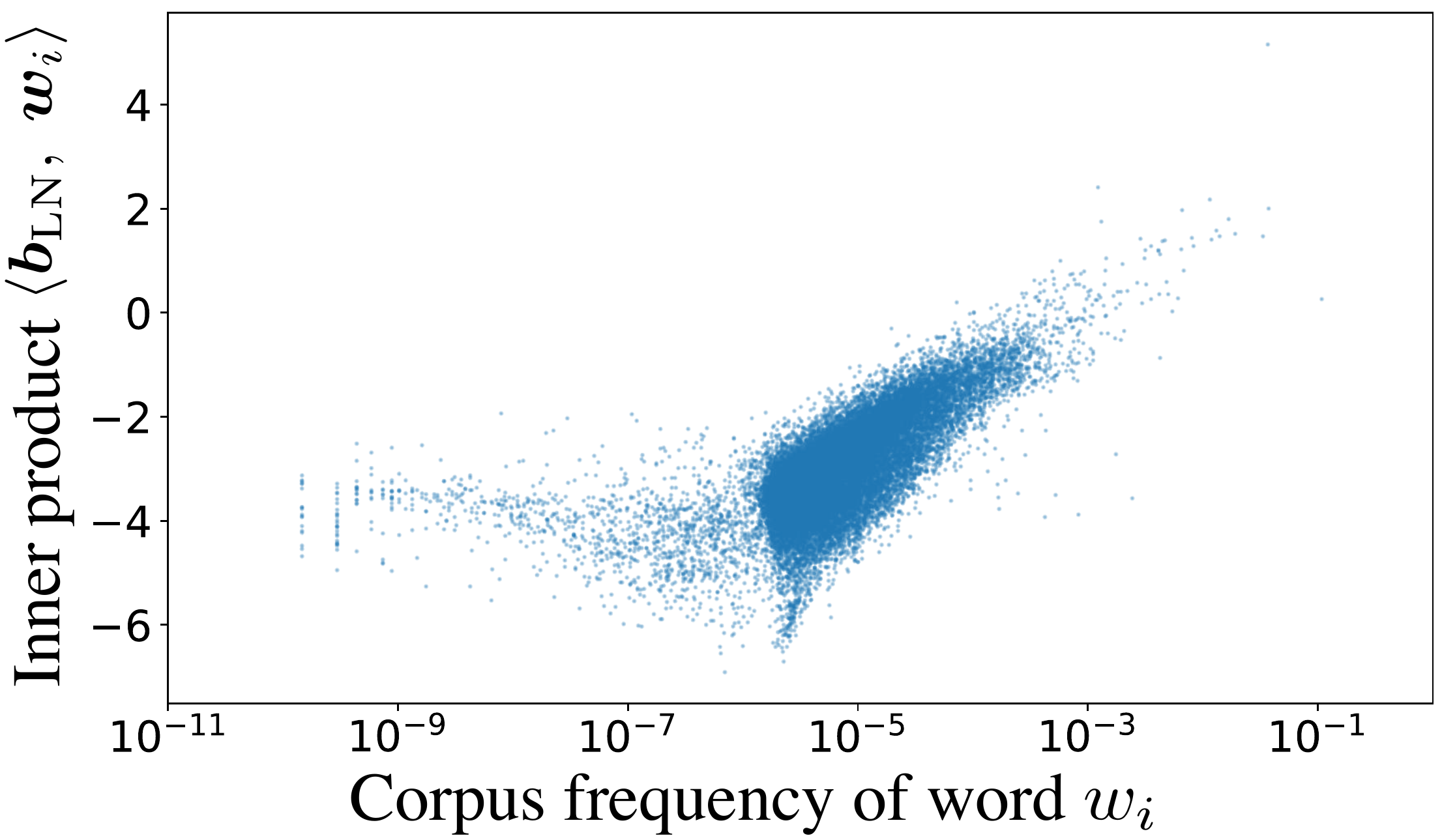}
\caption{
    Relationship between the corpus word frequency and the inner product of $\bm b_{\mathrm{LN}}$ and each output word embedding $\bm w_i$ in BERT large.
}
\label{fig:bert_large_bias_parameter}
\end{figure}

\begin{figure}[ht]
\centering
\includegraphics[width=\hsize]{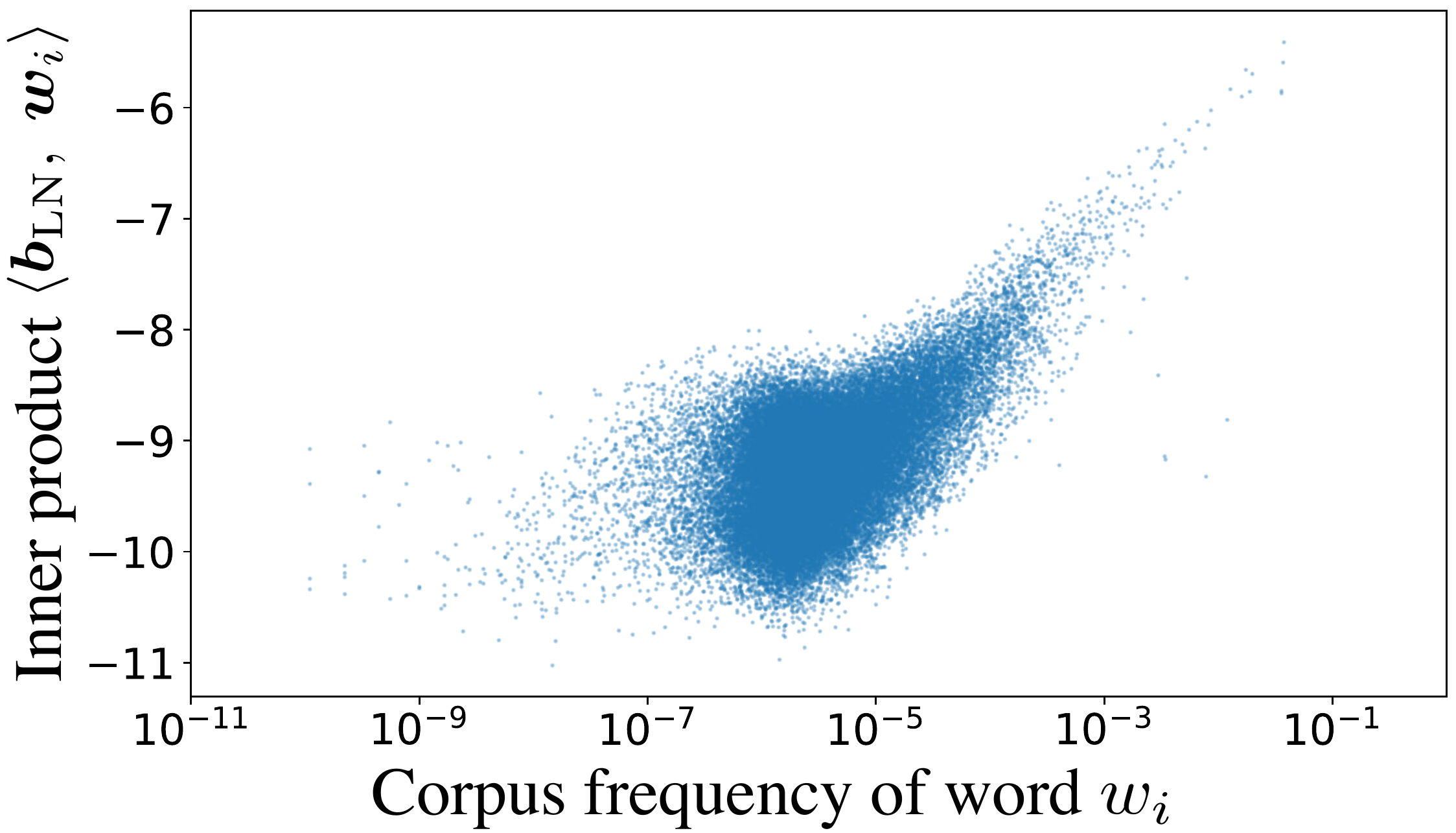}
\caption{
    Relationship between the corpus word frequency and the inner product of $\bm b_{\mathrm{LN}}$ and each output word embedding $\bm w_i$ in GPT-2 medium.
}
\label{fig:gpt2_medium_bias_parameter}
\end{figure}

\begin{figure}[ht]
\centering
\includegraphics[width=\hsize]{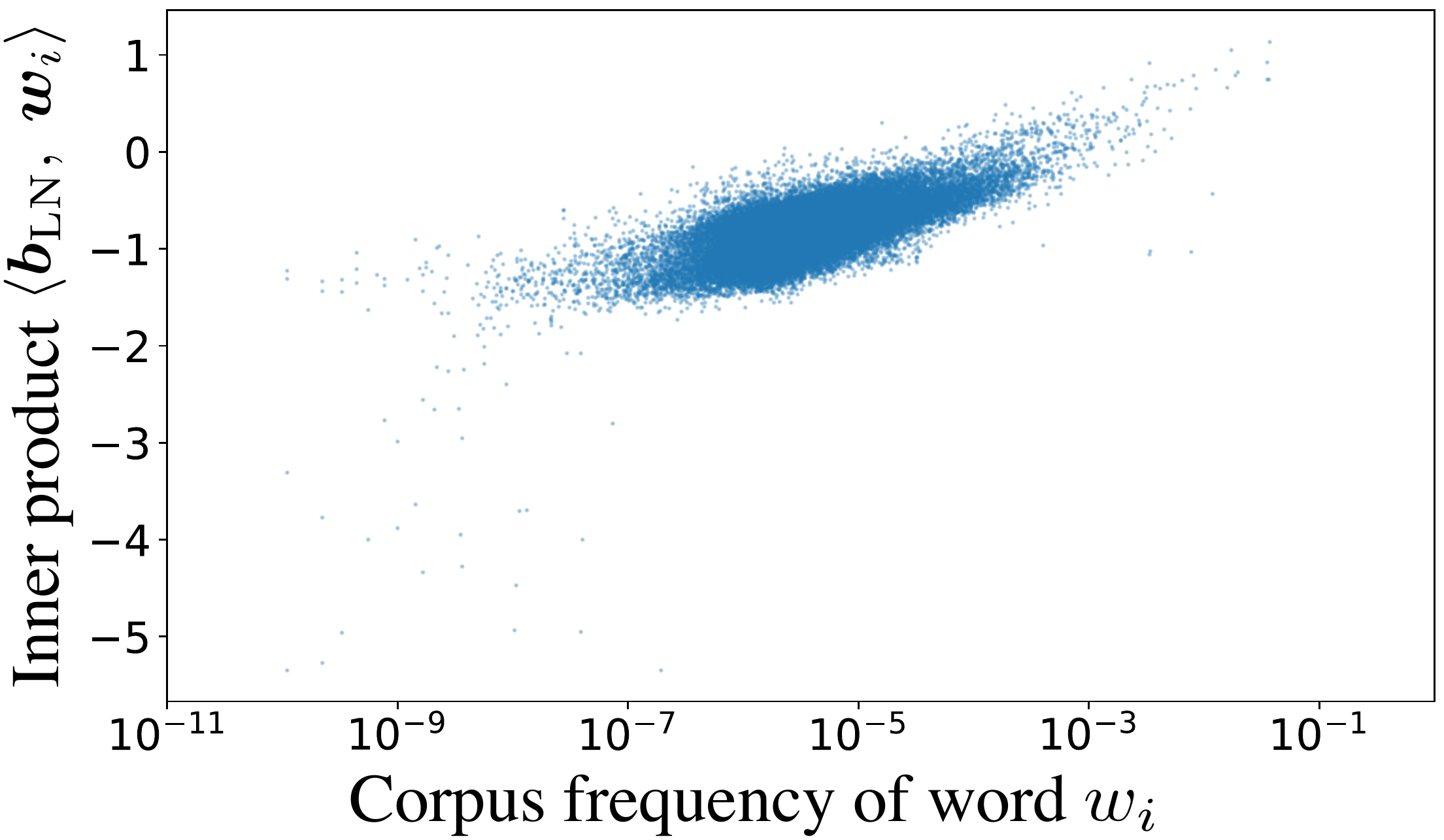}
\caption{
   Relationship between the corpus word frequency and the inner product of $\bm b_{\mathrm{LN}}$ and each output word embedding $\bm w_i$ in GPT-2 large.
}
\label{fig:gpt2_large_bias_parameter}
\end{figure}

\begin{figure}[ht]
\centering
\includegraphics[width=\hsize]{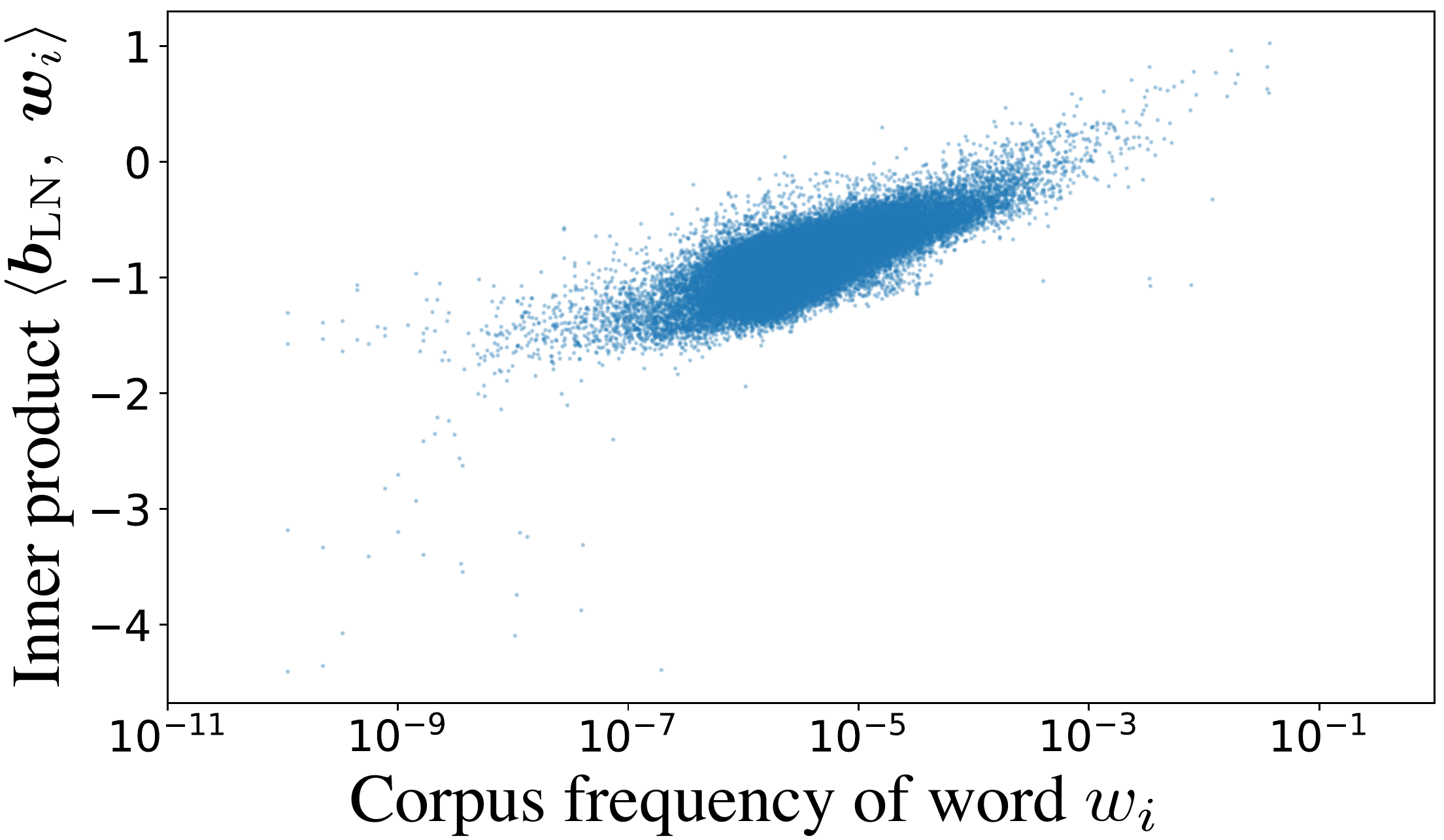}
\caption{
    Relationship between the corpus word frequency and the inner product of $\bm b_{\mathrm{LN}}$ and each output word embedding $\bm w_i$ in GPT-2 xl.
}
\label{fig:gpt2_xl_bias_parameter}
\end{figure}

In Section~\ref{sec:exp:geometrical_observation}, we showed that the inner product of $\bm b_{\mathrm{LN}}$ and each output word embedding $\bm w_i$ correlated well with the word frequency for GPT-2 small (Figure~\ref{fig:gpt2_bias_parameter}).
The results for the other models are shown in Figures~\ref{fig:bert_base_bias_parameter} to \ref{fig:gpt2_xl_bias_parameter}.
The Spearman's correlation coefficient is listed in Table~\ref{tab:spearman}.

In Section~\ref{sec:exp2:control_bias}, we showed the effect of controlling the bias $\bm b_{\mathrm{LN}}$ on GPT-2's text generation with a top-p sampling strategy.
We also conducted experiments with other sampling strategies: top-k sampling~\cite{Holtzman2020nucleussampling} and vanilla sampling.
The results of these two sampling strategies are listed in Tables~\ref{tab:generated_text_eval_topk} and~\ref{tab:generated_text_eval_full}.
We found that the results of top-k sampling were similar to those of top-p sampling; 
for the larger models, GPT-2 large ($\lambda=0.3$) and xl ($\lambda=0.5$), there also was a sweet spot, where diversity and \textsc{Mauve} improved with little decrease in \textsc{Ppl}.
In contrast, with vanilla sampling, both \textsc{Mauve} and \textsc{Ppl} decreased consistently and quickly.

\begin{table}[t]
\centering
\small
\begin{tabular}{@{}llc@{}}
\toprule
\multicolumn{2}{c}{Model}
& \multicolumn{1}{c}{Spearman's $\rho$} \\ 
\cmidrule{1-3}
\morecmidrules
\cmidrule{1-3}
\multirow{2}{*}{BERT}
& base   & 0.84 \\
& large  &  0.74 \\
\cmidrule{1-3}
\multirow{4}{*}{GPT-2}
& small  &  0.78  \\
& medium & 0.43  \\
& large  &  0.61   \\
& xl     &  0.70 \\ 
\bottomrule
\end{tabular}
\caption{
The Spearman's correlation coefficient between the corpus word frequency and inner product of $\bm b_{\mathrm{LN}}$ and each output word embedding $\bm w_i$.
}
\label{tab:spearman}
\end{table}

\begin{table}[t]
\centering
\small
\setlength{\tabcolsep}{5pt}
\begin{tabular}{@{}ccccccc@{}}
\toprule
\multicolumn{1}{c}{\multirow{2}{*}{Model}} & \multicolumn{1}{c}{\multirow{2}{*}{\hspace*{1.5mm}$\lambda$\hspace*{1.5mm}}} & \multicolumn{3}{c}{Diversity ↑} & \multicolumn{2}{c}{Quality} \\ 
\cmidrule(lr){3-5} \cmidrule(l){6-7} 
\multicolumn{1}{c}{} 
& \multicolumn{1}{c}{} 
& \multicolumn{1}{c}{$D_1$} 
& \multicolumn{1}{c}{$D_2$} 
& \multicolumn{1}{c}{$D$} 
& \multicolumn{1}{c}{\textsc{Mauve} ↑} 
& \multicolumn{1}{c}{\textsc{Ppl} ↓} \\ 
\cmidrule{1-7}
\morecmidrules
\cmidrule{1-7}
\multirow{4}{*}{small} 
 & $1$  &  0.03  & 0.23 &  0.42  &  0.78 & \textbf{19.4} \\
 & $0.9$ & 0.03 &  0.27 & 0.45 & \textbf{0.82} & 19.8  \\
 & $0.7$ & \textbf{0.03} &  \textbf{0.34} & \textbf{0.48} & 0.72 & 22.0  \\
 & $0$  & 0.02  & 0.12  & 0.13  & 0.01 & 65.9   \\ 
\cmidrule{1-7}
\multirow{3}{*}{med.}
 & $1$  &  0.03  &  0.27 &  0.46  &  \textbf{0.89} & \textbf{14.6}  \\
 & $0.3$ & 0.03 & \textbf{0.38}  & 0.50 & 0.64 & 17.8  \\
& $0$  & 0.03  & 0.33 & 0.46 & 0.22 & 21.3   \\
\cmidrule{1-7}
\multirow{3}{*}{large}
 & $1$  & 0.03 & 0.26  & 0.44 & 0.89  &  \textbf{12.7}   \\
 & $0.3$ & \textbf{0.03} & 0.32 & 0.48 & \textbf{0.90} & 13.1  \\
 & $0$  & 0.03 & \textbf{0.34} & \textbf{0.50}  & 0.87 & 13.6 \\ 
\cmidrule{1-7}
\multirow{3}{*}{xl}
 & $1$  & 0.03  & 0.28  &  0.45   &  0.92 &  \textbf{11.4}   \\
 & $0.5$ & 0.03 & 0.32 & 0.48 & \textbf{0.92} &  11.6 \\
 & $0$  & \textbf{0.03} & \textbf{0.36} & \textbf{0.50} & 0.89 & 12.1  \\ 
\bottomrule
\end{tabular}
\caption{
    Evaluation results for GPT-2 (top-k sampling) while bias $\bm b_{\mathrm{LN}}$ was controlled with $\lambda$.
}
\label{tab:generated_text_eval_topk}
\end{table}

\begin{table}[t]
\centering
\small
\setlength{\tabcolsep}{5pt}
\begin{tabular}{@{}ccccccc@{}}
\toprule
\multicolumn{1}{c}{\multirow{2}{*}{Model}} & \multicolumn{1}{c}{\multirow{2}{*}{\hspace*{1.5mm}$\lambda$\hspace*{1.5mm}}} & \multicolumn{3}{c}{Diversity ↑} & \multicolumn{2}{c}{Quality} \\ 
\cmidrule(lr){3-5} \cmidrule(l){6-7} 
\multicolumn{1}{c}{} 
& \multicolumn{1}{c}{} 
& \multicolumn{1}{c}{$D_1$} 
& \multicolumn{1}{c}{$D_2$} 
& \multicolumn{1}{c}{$D$} 
& \multicolumn{1}{c}{\textsc{Mauve} ↑} 
& \multicolumn{1}{c}{\textsc{Ppl} ↓} \\ 
\cmidrule{1-7}
\morecmidrules
\cmidrule{1-7}
\multirow{3}{*}{small} 
 & $1$  &   0.07 &   0.49  & 0.59 &  \textbf{0.50} & \textbf{19.4} \\
 & $0.5$ &   \textbf{0.14}  &  \textbf{0.88}   &  0.73  & 0.02  &   27.0   \\
 & $0$    & 0.12  &  0.71 &  0.61   &  0.01 & 65.9   \\ 
\cmidrule{1-7}
\multirow{3}{*}{med.}
 & $1$  &  0.09  & 0.56 & 0.63  &  \textbf{0.33} & \textbf{14.6}  \\
 & $0.2$  &  0.19  &  \textbf{0.86}   & \textbf{0.74}  &   0.03   &  18.8    \\
& $0$  &  \textbf{0.21}  &  0.86   & 0.74  &   0.02   &  21.3   \\
\cmidrule{1-7}
\multirow{3}{*}{large}
 & $1$  &  0.06   &  0.44  &  0.56  &   \textbf{0.77}  &  \textbf{12.7}   \\
 & $0.5$  &  0.08  &  0.55  &   0.61  &  0.53   &  12.9   \\
 & $0$   &   \textbf{0.11}   &  \textbf{0.69}  & \textbf{0.67}  &  0.22  &  13.6 \\ 
\cmidrule{1-7}
\multirow{3}{*}{xl}
 & $1$  &  0.06   &  0.43  &  0.56  &  \textbf{0.82}  &  \textbf{11.4}   \\
 & $0.5$  &  0.08  &  0.54  &  0.61   &  0.61   &   11.6   \\
 & $0$  & \textbf{0.11}  &  \textbf{0.68}  &  \textbf{0.67}  &   0.24  &  12.1  \\ 
\bottomrule
\end{tabular}
\caption{
    Evaluation results for GPT-2 (vanilla sampling) while bias $\bm b_{\mathrm{LN}}$ was controlled with $\lambda$.
}
\label{tab:generated_text_eval_full}
\end{table}

\clearpage
\begin{table*}[ht]
\centering
\setlength{\tabcolsep}{3.5pt}  %
\small
\begin{tabular}{@{}lcl@{}}
\toprule
Model &
\multicolumn{1}{c}{$\lambda$} & \multicolumn{1}{c}{Generated text} \\ 
\cmidrule{1-3}
\morecmidrules
\cmidrule{1-3}
\multirow{11}{*}{small} 
& 1 &  
\begin{tabular}{l} There has been one product that I've wanted for a while --- that is baseball's fountain and. I wanted to try to \\ get another product to make it as polished and simple to use and even easier to push the right buttons. Have \\ you played with some of the furniture brands of the past? Do you think the new smart building is going to... \end{tabular}\\
\cmidrule{2-3}
& 0.6 &  
\begin{tabular}{l} There has been one product that I've wanted for awhile: \textbf{Asus ZenUI} Keyboard Replacement Kit FAQ. I \\ purchased this replacement keyboard replacement kit prior to 2014 when \textbf{Asus} shipped its \textbf{ZenUI} \\ \multicolumn{1}{c}{...} \\ BIOS Reset Warranty Long warranty \underline{EUR 3500 EUR 4550 EUR 470 EUR 520 EUR 590 EUR 560 EUR}...\end{tabular}\\
\cmidrule{2-3}
& 0 &  
\begin{tabular}{l} There has been one product that I've wanted for awhile got released that alters baseball's bench press. I \\ mention \textbf{Alejandro Nazarovski} prior thus preferring \textbf{Julian Whitaker} however altering \textbf{Alejandro} \\ \multicolumn{1}{c}{...} \\ combined with dumbbell movements \uwave{combined with negatives ratios Improved athlete mobility Decreased} \\ \uwave{fatigue Diseases Whilst adjusting lifts Underestimating injury Potential Extensions Suspension Period}... \end{tabular}\\
\cmidrule{1-3}
\multirow{8}{*}{large} 
& 1 & 
\begin{tabular}{l} The \textbf{Atlanta Falcons} have started the 2015 season 4-0. (Photo: \textbf{Winslow Townson} / \textbf{Associated Press}) The \\ Falcons' longest streak of consecutive seasons with a winning record started on the same day in Week 11 that \\ \textbf{Mike Shanahan} and the Falcons experienced one of their most compelling victories of the season...\end{tabular}\\
\cmidrule{2-3}
& 0.5 & 
\begin{tabular}{l} The \textbf{Atlanta Falcons} have started the 2015 season 4-1, including a triumph over the \textbf{New Orleans Saints} at \\ \textbf{Mercedes-Benz Stadium}. Look at what this team could be capable of as the season progresses. It has the \\ goods, the direction, the talent to make a run at becoming a legitimate \textbf{Super Bowl} contender. More...\end{tabular}\\
\cmidrule{2-3}
& 0 & 
\begin{tabular}{l} The \textbf{Atlanta Falcons} have started the 2015 season 4-0, including a win over the \textbf{Minnesota Vikings} last \\ Sunday night. It's been a perfect start to 2014 as well. Looking ahead, what's the road ahead?\\
Week 1 @ \textbf{Tampa Bay Buccaneers} ... \end{tabular}\\
\bottomrule
\end{tabular}
\caption{
Examples of text generated by GPT-2 small and large with top-p sampling while bias $\bm b_{\mathrm{LN}}$ was controlled with $\lambda$.
Proper nouns are in \textbf{bold}, repetitions of similar phrases are \uline{straight underlined}, and ungrammatical passages are highlighted with \uwave{wavy underlines}.
Note that the first 10 words are given to the model as context.
}
\label{tab:example_small}
\end{table*}

\section{Detailed experimental settings}
\label{ap:generate_detail}
To observe the TLM word prediction distribution (the main experiments in Section~\ref{sec:exp1:analyze_bias} and the measurement of \textsc{Ppl} in Section~\ref{sec:exp2:control_bias}), we let BERT predict words corresponding to \texttt{[MASK]} tokens, and we let GPT-2 predict the second and subsequent words in each sequence.
If the length of an input sequence was greater than the maximum input length $k$ of the model, only the first $k$ words were used.

To evaluate the TLM text generation (Section~\ref{sec:exp2:control_bias}), the first 10 words of each sequence were fed into to GPT-2, and subsequent words were generated until the length of the sequence reached 1,024 words or the end-of-sequence token was generated.
For GPT-2 small and medium, we varied $\lambda$ in increments of $0.1$ to control the bias $\bm b_{\mathrm{LN}}$.
For GPT-2 large and xl, we first checked the results for 100 samples and obtained the values with some kind of trends; we then varied $\lambda$ in $\{0, 0.3, 0.5, 0.7, 1.0\}$ for the entire dataset, including the values.

We experimented with three decoding strategies: vanilla sampling, top-k sampling, and top-p sampling.
In the top-k sampling, $k$ was set to $50$.
In the top-p sampling, $p$ was set to $0.9$.
Furthermore, before we evaluated the model-generated texts with the N-gram based diversity metrics, we applied the word tokenizer provided by NLTK~\cite{bird-loper-2004-nltk}.

\section{Examples of generated text}
\label{ap:examples_generated_text}

Table~\ref{tab:example_small} shows examples of text generated by GPT-2 small and large while controlling the bias $\bm b_{\mathrm{LN}}$ with $\lambda$.

\end{document}